\documentclass{article}

\usepackage[preprint]{neurips_2019}

\usepackage[utf8]{inputenc} 
\usepackage[T1]{fontenc}    
\usepackage[hidelinks]{hyperref}       
\usepackage{url}            
\usepackage{booktabs}       
\usepackage{amsfonts}       
\usepackage{nicefrac}       
\usepackage{microtype}      

\usepackage{smile}
\usepackage{color}
\usepackage{xcolor}

\newcommand{\shortsection}[1]{\vspace*{1ex}\noindent{\bf #1.}}
\def \Risk {\text{Risk}}
\def \AdvRisk {\text{AdvRisk}}
\def \Ball {\cB}
\def \Rob {\text{Rob}}

\makeatletter
\renewcommand*{\@fnsymbol}[1]{\ifcase#1\or*\else\@arabic{\numexpr#1-1\relax}\fi}
\makeatother

\definecolor{carnelian}{rgb}{0.7, 0.11, 0.11}
\newcommand{\ifcomments}{\iftrue}

\ifx\condtion\undefined
\newtheorem{condition}[theorem]{Condition}
\fi

\title{Understanding the Intrinsic Robustness of Image Distributions using Conditional Generative Models}

\author{ 
Xiao Zhang\thanks{Equal Contribution} \\ \small University of Virginia \\ \small \textsf{xz7bc@virginia.edu} \And Jinghui Chen$^{*}$ \\ \small University of California, \\ \small Los Angeles \\ \small \textsf{jinghuic@ucla.edu} \And Quanquan Gu \\ \small University of California, \\ \small Los Angeles \\ \small \textsf{qgu@cs.ucla.edu} \And David Evans \\ \small University of Virginia \\ \small \textsf{evans@virginia.edu}
}

\begin{document}

\maketitle

\begin{abstract}
Starting with \citet{gilmer2018adversarial}, several works have demonstrated the inevitability of adversarial examples based on different assumptions about the underlying input probability space.  It remains unclear, however, whether these results apply to natural image distributions. In this work, we assume the underlying data distribution is captured by some conditional generative model, and prove intrinsic robustness bounds for a general class of classifiers, which solves an open problem in \citet{fawzi2018adversarial}. Building upon the state-of-the-art conditional generative models, we study the intrinsic robustness of two common image benchmarks under $\ell_2$ perturbations, and show the existence of a large gap between the robustness limits implied by our theory and the adversarial robustness achieved by current state-of-the-art robust models. Code for all our experiments is available at {\small\url{https://github.com/xiaozhanguva/Intrinsic-Rob}}.
\end{abstract}

\section{Introduction}
\label{sec:intro}

Deep neural networks (DNNs) have achieved remarkable performance on many visual \citep{sutskever2012imagenet,he2016deep} and speech \citep{hinton2012deep} recognition tasks, but recent studies have shown that state-of-the-art DNNs are surprisingly vulnerable to \emph{adversarial perturbations}, small imperceptible input transformations that are designed to switch the prediction of the classifier \citep{szegedy2014intriguing, goodfellow2015explaining}. This 
has led to 
a vigorous arms race between heuristic defenses \citep{papernot2016distillation,madry2017towards, chakraborty2018adversarial, wang2019convergence} that propose ways to defend against existing attacks and newly-devised attacks \citep{carlini2017towards,athalye2018obfuscated,tramer2020adaptive}
that are able to penetrate such defenses. Reliable defenses appear to be elusive, despite progress on provable defenses, including formal verification  \citep{katz2017reluplex, tjeng2018evaluating}  and relaxation-based certification methods \citep{sinha2017certifying,raghunathan2018certified,wong2018provable,
gowal2018effectiveness, wang2018mixtrain}. Even the strongest of these defenses leave large opportunities for adversaries to find adversarial examples, while suffering from high computation costs and scalability issues.

Witnessing the difficulties of constructing robust classifiers, a line of recent works \citep{gilmer2018adversarial,fawzi2018adversarial,mahloujifar2018curse,shafahi2018adversarial} aims to understand the limitations of robust learning by providing theoretical bounds on adversarial robustness for arbitrary classifiers. By imposing different assumptions on the underlying data distributions and allowable perturbations, all of these theoretical works show that no adversarially robust classifiers exist for an assumed metric probability space, as long as the perturbation strength is sublinear in the typical norm of the inputs. Although such impossibility results seem disheartening to the goal of building robust classifiers, it remains unknown to what extent real image distributions satisfy the assumptions  needed to obtain these results.

In this paper, we aim to bridge the gap between the theoretical robustness analyses on well-behaved data distributions and the maximum achievable adversarial robustness, which we call \emph{intrinsic robustness} (formally defined by Definition \ref{def:intrinsic robustness}), for typical image distributions. 
More specifically, we assume the underlying data lie on a separable low-dimensional manifold, which can be captured using a conditional generative model, then systematically study the intrinsic robustness based on the conditional generating process from both theoretical and experimental perspectives. Our main contributions are:

\begin{itemize}
\item We prove a fundamental bound on intrinsic robustness (Section \ref{sec:theory}), provided that the underlying data distribution can be captured by a conditional generative model, solving an open problem in \citet{fawzi2018adversarial}.
    
\item Building upon a trained conditional generative model that mimics the underlying data generating process, we empirically  evaluate the intrinsic robustness on image distributions based on MNIST and ImageNet (Section \ref{sec:exp-lip}). Our estimates of intrinsic robustness demonstrate that there is still a large gap between the limits implied by our theory and the state-of-the-art robustness achieved by robust training methods (Section \ref{sec:exp comp}).
    
\item We theoretically characterize the fundamental relationship between the \emph{in-distribution adversarial risk} (which restricts adversarial examples to lie on the image manifold, and is formally defined by Definition \ref{def:in dist adv risk}) and the intrinsic robustness (Remark \ref{rem:relation}), and propose an optimization method to search for in-distribution adversarial examples with respect to a given classifier. 
Our estimated in-distribution robustness for state-of-the-art adversarially trained classifiers, together with the derived intrinsic robustness bound, provide a better understanding on the intrinsic robustness for natural image distributions (Section~\ref{sec:exp-rob}). 
\end{itemize}

\shortsection{Notation}
We use lower boldfaced letters such as $\bx$ to denote vectors, and $[n]$ to denote the index set $\{1,2,\ldots,n \}$. 
For any $\bx\in\cX$ and $\epsilon\geq 0$, denote by $\cB(\bx,\epsilon,\Delta)=\{\bx'\in\cX:\Delta(\bx,\bx')\leq\epsilon\}$ the $\epsilon$-ball around $\bx$ with radius $\epsilon$ in some distance metric $\Delta$.
When the metric is free of context, we simply write $\cB(\bx,\epsilon) = \cB(\bx,\epsilon,\Delta)$.
We use $\cN(\zero,\bI_d)$ to denote the $d$-dimensional standard Gaussian distribution, and let $\nu_d$ be its probability measure. 
For the one dimensional case, we use $\Phi(x)$ to denote the cumulative distribution function (CDF) of $\cN(0,1)$, and use $\Phi^{-1}(x)$ to denote its inverse function. 
For any function $g:\cZ\rightarrow\cX$ and probability measure $\nu$ defined over $\cZ$, $g_*(\nu)$ denotes the push-forward measure of $\nu$. The $\ell_2$-norm of a vector $\bx\in\RR^n$ is defined as $\|\bx\|_{2} = (\sum_{i\in[n]} x_i^2)^{1/2}$.

\section{Related Work}
\label{sec:related}

Several recent works \citep{gilmer2018adversarial,mahloujifar2018curse,shafahi2018adversarial,dohmatob2018generalized,bhagoji2019lower} derived theoretical bounds on maximum achievable adversarial robustness using isoperimetric inequality under different assumptions of the input space. For instance, 
based on the assumption that the input data are uniformly distributed over two concentric $n$-spheres \citep{gilmer2018adversarial} or the underlying metric probability space satisfies a concentrated property \citep{mahloujifar2018curse}, any classifier with constant test error was proven to be vulnerable to adversarial perturbations sublinear to the input dimension. 
\citet{shafahi2018adversarial} showed that adversarial examples are inevitable, provided the maximum density of the underlying input distribution is small relative to uniform density. However, none of the above theoretical works provide any experiments to justify the imposed assumptions hold for real datasets, thus it is unclear whether the derived theoretical bounds are meaningful for typical image distributions. Our work belongs to this line of research, but encompasses the practical goal of understanding the robustness limits for real image distributions.

The most related literature to ours is \citet{fawzi2018adversarial}, which proved a classifier-independent upper bound on intrinsic robustness, provided the underlying distribution is well captured by a smoothed generative model with Gaussian latent space and small Lipschitz parameter.
However, their proposed theory cannot be applied to image distributions that lie on a low-dimensional, non-smooth manifold, as their framework requires examples from different classes to be close enough in the latent space. 
In contrast, our proposed theoretical bounds on intrinsic robustness are more general in that they can be applied to non-smoothed data manifolds, such as image distributions generated by conditional models. In addition, we propose an empirical method to estimate the intrinsic robustness on the generated image distributions under worst-case $\ell_2$ perturbations.

\citet{mahloujifar2019empirically} proposed to understand the inherent limitations of robust learning using heuristic methods to measure the concentration of measure based on a given set of i.i.d. samples. However, it is unclear to what extent the estimated sample-based concentration approximates the actual intrinsic robustness with respect to the underlying data distribution. In comparison, we assume the underlying data distribution can be captured by a conditional generative model and directly study the robustness limit on the generated data distribution.
\section{Preliminaries}
\label{sec:prob}
We focus on the task of image classification.
Let $(\cX,\mu, \Delta)$ be a metric probability space, where $\cX\subseteq\RR^n$ denotes the input space, $\mu$ is a probability distribution over $\cX$ and $\Delta$ is some distance metric defined on $\cX$. Suppose there exists a ground-truth function, $f^*:\cX\rightarrow[K]$, that gives a label to any image $\bx\in\cX$, where $[K]$ denotes the set of all possible class labels. 
The objective of classification is to learn a function $f:\cX\rightarrow [K]$ that approximates $f^*$ well. In the context of adversarial examples, $f$ is typically evaluated based on \emph{risk}, which captures the classification accuracy of $f$ on normal examples, and \emph{adversarial risk}, which captures the classifier's robustness against adversarial perturbations:

\begin{definition}
\label{def:risk and adv risk}
Let $(\cX,\mu,\Delta)$ be a metric probability space and $f^*$ be the ground-truth classifier. For any classifier $f$, the \emph{risk} of $f$ is defined as:
\begin{align*}
    \Risk_\mu(f) = \Pr_{\bx\sim\mu} \big[f(\bx) \neq f^*(\bx) \big].
\end{align*}
The \emph{adversarial risk} of $f$ against perturbations with strength $\epsilon$ in metric $\Delta$ is defined as:
\begin{align*}
&\AdvRisk_\mu^\epsilon(f) = \Pr_{\bx\sim\mu} \big[\exists\: \bx'\in\Ball(\bx,\epsilon) \:\text{ s.t. } f(\bx')\neq f^*(\bx')\big].
\end{align*}
\end{definition}

Other definitions of adversarial risk also exist in literature, such as the definition used in \citet{madry2017towards} and the one proposed in \citet{fawzi2018adversarial}. However, these definitions are equivalent to each other under the assumption that small perturbations do not change the ground-truth labels.
Another closely-related definition for adversarial robustness is the expected distance to the nearest error (see \citet{diochnos2018adversarial} for the relation between these definitions). Our results can be applied to this definition as well.

Under different assumptions of the input metric probability space, previous works proved model-independent bounds on adversarial robustness. \emph{Intrinsic robustness}, defined originally by \citet{mahloujifar2019empirically}, captures the maximum adversarial robustness that can be achieved for a given robust learning problem: 

\begin{definition}
\label{def:intrinsic robustness} 
Using the same settings as in Definition \ref{def:risk and adv risk} and let $\cF$ be some class of classifiers. The \emph{intrinsic robustness} with respect to $\cF$ is defined as:
\begin{equation*}
\mathrm{Rob}_\mu^{\epsilon}(\cF) = 1-\inf_{f\in\cF}\big\{\AdvRisk_\mu^\epsilon(f) \big\}.
\end{equation*}
In this work, we consider the class of imperfect classifiers that have risk at least some $\alpha>0$.
\end{definition}

Motivated by the great success of producing natural-looking
images using conditional generative adversarial nets (GANs)~\citep{mirza2014conditional,odena2017conditional,brock2018large}, we assume the underlying data distribution $\mu$ can be modeled by some \emph{conditional generative model}. A generative model can be seen as a function $g:\cZ\rightarrow \cX$ that maps some latent distribution, usually assumed to be multivariate Gaussian, to some generated distribution over $\cX$.
 
Conditional generative models incorporate the additional class information into the data generating process.
A conditional generative model can be considered as a set of generative models $\{g_i\}_{i\in[K]}$, where images from the $i$-th class can be generated by transforming latent Gaussian vectors through $g_i$. 
More rigorously, we say a probability distribution $\mu$ can be generated by a conditional generative model $\{(g_i,p_i)\}_{i\in[K]}$, if $\mu=\sum_{i=1}^K p_i\cdot (g_i)_*(\nu_d)$, where $K$ is the total number of different class labels, and $p_i\in[0,1]$ represents the probability of sampling an image from class $i$. 

Based on the conditional model, we introduce the definition of \emph{in-distribution adversarial risk}:

\begin{definition}
\label{def:in dist adv risk}
Consider the same settings as in Definition \ref{def:risk and adv risk}. Suppose $\mu$ can be captured by a conditional generative model $\{(g_i,p_i)\}_{i\in[K]}$. For any given classifier $f$,  the \emph{in-distribution adversarial risk} of $f$ against $\epsilon$-perturbations is defined as:
\begin{align*}
    \text{In-AdvRisk}_\mu^\epsilon(f) = \Pr_{(\bx,i)\sim\mu}\big[ \exists\: \bz'\in\cZ \:\text{ s.t. } g_i(\bz') \in \Ball(\bx,\epsilon) 
    \text{ and } f(g_i(\bz'))\neq f^*(g_i(\bz'))\big].
\end{align*}
\end{definition}

Given the fact that the in-distribution adversarial risk restricts the adversarial examples to be on the image manifold, it holds that, for any classifier $f$, $\text{In-AdvRisk}_\mu^\epsilon(f)\leq \AdvRisk_\mu^\epsilon(f)$. As will be shown in the next section, such a notion of in-distribution adversarial risk is closely related to the intrinsic robustness for the considered class of imperfect classifiers.
\section{Main Theoretical Results}
\label{sec:theory}

In this section, we present our main theoretical results on intrinsic robustness, provided the underlying distribution can be modeled by some conditional generative model (our results and proof techniques could also be easily applied to unconditional generative models).  Based on the underlying generative process, the following local Lipschitz condition connects perturbations in the image space to the latent space. 

\begin{condition} 
\label{def:lipschitz}
Let $g:\RR^d\rightarrow\cX$ be a generative model that maps the latent Gaussian distribution $\nu_d$ to some generated distribution. Consider Euclidean distance as the distance metric for $\RR^d$, and $\Delta$ as the metric for $\cX$. 
Given $r>0$, $g$ is said to be $L(r)$-locally Lipschitz with probability at least $1-\delta$, if it satisfies
\begin{align*}
    \Pr_{\bz\sim\nu_d}\Big[\forall \bz'\in\Ball(\bz,r),\: \Delta\big(g(\bz'),g(\bz)\big) \leq L(r)\cdot\|\bz'-\bz\|_2  \Big] \geq 1-\delta.
\end{align*}
\end{condition}

As the main tool for bounding the intrinsic robustness, we present the Gaussian Isoperimetric inequality for the sake of completeness. This inequality, proved by \citet{borell1975brunn} and \citet{sudakov1978extremal}, bounds the minimum expansion of any subset with respect to the standard Gaussian measure.

\begin{lemma}[Gaussian Isoperimetric Inequality]\label{lem:Gaussian isoperimetric inq}
Consider metric probability space $(\RR^d,\nu_d,\|\cdot\|_2)$, where $\nu_d$ is the probability measure for $d$-dimensional standard Gaussian distribution $\cN(\zero,\bI_d)$, and $\|\cdot\|_2$ denotes the Euclidean distance. For any subset $\cE\subseteq\RR^d$ and $r\geq 0$, let $\cE_r = \big\{\bz\in\RR^d: \exists \bz'\in\cE, \text{ s.t. } \|\bz-\bz'\|_2\leq r\big\}$ be the $r$-expansion of $\cE$, then it holds that
\begin{equation}
\label{eq:gaussian isoperimentric inequality}
    \nu_d(\cE_r) \geq \Phi\big(\Phi^{-1}\big(\nu_d(\cE)\big)+r\big),
\end{equation}
where $\Phi(x) = \frac{1}{\sqrt{2\pi}}\int_{-\infty}^x \exp(-u^2/2) du$ is the CDF of $\cN(0,1)$, and $\Phi^{-1}(x)$ denotes its inverse.
\end{lemma}

In particular, when $\cE$ belongs to the set of half-spaces, the equality is achieved in \eqref{eq:gaussian isoperimentric inequality}. 

Making use of the Gaussian Isoperimetric Inequality and the local Lipschitz condition of the conditional generator, the following theorem proves a lower bound on the (in-distribution) adversarial risk for any given classifier, provided the underlying distribution can be captured by a conditional generative model. 

\begin{theorem}\label{thm:AdvRisk lower bound}
Let $(\cX, \mu, \Delta)$ be a metric probability space and $f^*:\cX\rightarrow [K]$ be the underlying ground-truth. Suppose $\mu$ can be generated by a conditional generative model $\{(g_i,p_i)\}_{i\in [K]}$.
Given $\epsilon>0$, suppose there exist constants $r>0$ and $\delta\in(0,1]$ such that for any $i\in[K]$, $g_i$ satisfies $L_i(r)$-local Lipschitz property with probability at least $1-\delta$ and $r\cdot L_i(r)\geq \epsilon$. Then for any classifier $f$, it holds that
\begin{align*}  
\AdvRisk_\mu^\epsilon(f)\geq\text{In-AdvRisk}_\mu^\epsilon(f) \geq \sum_{i=1}^K p_i\cdot\Phi\bigg(\Phi^{-1}\big(\Risk_{\mu_i}(f)\big)+ \frac{\epsilon}{L_i(r)} \bigg) - \delta,
\end{align*}
where $\mu_i=(g_i)_*(\nu_d)$ is the pushforward measure of $\nu_d$ though $g_i$, for any $i\in[K]$.
\end{theorem}

We provide a proof in Appendix \ref{sec:proof 1}. Theorem \ref{thm:AdvRisk lower bound} suggests the (in-distribution) adversarial risk is related to the risk on each data manifold and the ratio between the perturbation strength and the Lipschitz constant. 

The following theorem, proved in Appendix \ref{sec:proof 2}, gives a theoretical upper bound on the intrinsic robustness with respect to the class of imperfect classifiers.

\begin{theorem}
\label{thm:intrinsic robustness bound}
Under the same setting as in Theorem \ref{thm:AdvRisk lower bound}, let $L_{\max}(r) = \max_{i\in[K]}L_i(r)$. Consider the class of imperfect classifiers $\cF_\alpha = \{f:\Risk_\mu(f)\geq\alpha\}$ with $\alpha > 0$, then the intrinsic robustness with respect to $\cF_\alpha$ can be bounded as,
\begin{align*}
    \Rob_\mu^\epsilon (\cF_\alpha) \leq 1+\delta - \min_{i\in[K]}\bigg\{p_i\cdot\Phi\bigg(\Phi^{-1}\bigg(\frac{\alpha}{p_i}\bigg)+ \frac{\epsilon}{L_{\max}(r)} \bigg)\bigg\},
\end{align*}
provided that $\alpha/p_i\leq 1$ for any $i\in[K]$. In addition, if we consider the family of classifiers that have conditional risk at least $\alpha$ for each class, namely $\tilde\cF_\alpha = \{f: \Risk_{\mu_i}(f)\geq\alpha, \forall i\in[K]\}$, then the intrinsic robustness with respect to $\tilde\cF_{\alpha}$ can be bounded by
\begin{align*}
    \Rob_\mu^\epsilon(\tilde\cF_\alpha) \leq 1+\delta - \sum_{i=1}^K p_i\cdot\Phi\bigg(\Phi^{-1}\big(\alpha\big)+ \frac{\epsilon}{L_{\max}(r)} \bigg).
    \end{align*}
\end{theorem}

\begin{remark}
Theorem \ref{thm:intrinsic robustness bound} shows that if the data distribution can be captured by a conditional generative model, the intrinsic robustness bound with respect to imperfect classifiers will largely depend on the ratio $\epsilon/L_{\max}$. For instance, if we assume the ratio $\epsilon/L_{\max} = 1$, then Theorem \ref{thm:intrinsic robustness bound} suggests that no classifier with initial risk at least $5\%$ can achieve robust accuracy exceeding $75\%$ for the assumed data generating process. In addition, if we assume the local Lipschitz parameter $L_{\max}$ is some constant, then adversarial robustness is indeed not achievable for high-dimensional data distributions, provided the perturbation strength $\epsilon$ is sublinear to the input dimension, which is the typical setting considered. 
\end{remark}

\begin{remark}
\label{rem:relation}
The intrinsic robustness is closely related to the in-distribution adversarial risk.
For the class of classifiers $\cF_\alpha$, one can prove that the intrinsic robustness is equivalent to the maximum achievable in-distribution adversarial robustness:
\begin{align}
\label{eq:equvilence}
\Rob_\mu^\epsilon(\cF_\alpha) = 1-\inf_{f\in\cF_\alpha}\{\text{In-AdvRisk}_\mu^\epsilon(f)\}.
\end{align}
Trivially, $\AdvRisk_\mu^\epsilon(f)\geq\text{In-AdvRisk}_\mu^{\epsilon}(f)$ holds for any $f$. For a given $f\in\cF_\alpha$, one can construct an $h_f\in\cF_\alpha$ such that $h_f(\bx)=f(\bx)$ if $\bx\in\cE_f\cap\cM$ and $h_f(\bx)=f^*(\bx)$ otherwise, where $\cE_f=\{\bx\in\cX: f(\bx)\neq f^*(\bx)\}$ denotes the error region of $f$ and $\cM$ is the considered image manifold. The construction immediately suggests $\text{In-AdvRisk}_\mu^\epsilon(f) = \text{AdvRisk}_\mu^\epsilon(h_f)$, which implies,
\begin{align*}
    \inf_{f\in\cF_\alpha}\{\text{In-AdvRisk}_\mu^\epsilon(f)\}= \inf_{f\in\cF_\alpha}\{\text{AdvRisk}_\mu^\epsilon(h_f)\} \geq \inf_{f\in\cF_\alpha}\{\text{AdvRisk}_\mu^\epsilon(f)\}.
\end{align*}
Combining both directions proves the soundness of \eqref{eq:equvilence}.
This equivalence suggests the in-distribution adversarial robustness of any classifier in $\cF_\alpha$ can be viewed as a lower bound on the actual intrinsic robustness, which motivates us to study the intrinsic robustness by estimating the in-distribution adversarial robustness of trained robust models in our experiments.
\end{remark}

\section{Experiments}
\label{sec:experiments}
This section provides our empirical evaluations of the intrinsic robustness on real image distributions to evaluate the tightness of our bound. We test our bound on two image distributions generated using MNIST \citep{lecun1998gradient} and ImageNet \citep{imagenet_cvpr09} datasets.

\subsection{Conditional GAN Models}
\begin{figure*}[bt]
    \centering
    \subfigure[ACGAN Generated MNIST]{\includegraphics[width=0.4\textwidth]{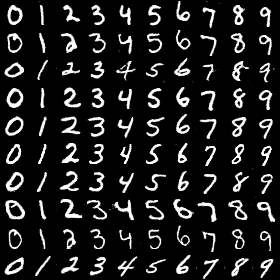}}
    \hspace{0.5in}
    \subfigure[BigGAN Generated ImageNet]{\includegraphics[width=0.4\textwidth]{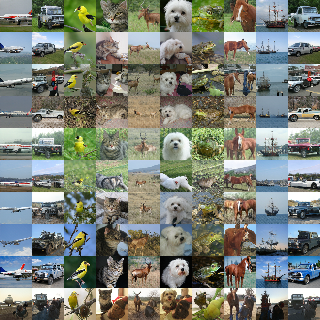}}
    \caption{Illustration of the generated images using different conditional models. For BigGAN generated images, we select $10$ specific classes from the $1000$ ImageNet classes (corresponding to the 10 image classes in CIFAR-10).
}
    \label{fig:generated images}
\end{figure*}

Instead of directly evaluating the robustness on real datasets, we make use of conditional GAN models to generate datasets from the learned data distributions and evaluate the robustness of several state-of-the-art robust models trained on the generated dataset for a fair comparison with the theoretical robustness limits.
Note that this approach is only feasible with conditional generative models as unconditional models cannot provide the corresponding labels for the generated data samples. 
For MNIST, we adopt ACGAN \citep{odena2017conditional} which features an additional auxiliary classifier for better conditional image generation. 
The ACGAN model generates $28 \times 28$ images from a $100$-dimension latent space concatenated with an addition $10$-dimension one-hot encoding of the conditional class labels. 
For ImageNet, we adopt the BigGAN model \citep{brock2018large} which is the state-of-the-art GAN model in conditional image generation. It generates $128 \times 128$ images from a $120$-dimension latent space. We down-sampled the generated images to $32 \times 32$ for efficiency propose. We consider a standard Gaussian\footnote{The original BigGAN model uses truncated Gaussian. We adapted it to standard Gaussian distribution.} as the latent distribution for both conditional generative models. 
Figure \ref{fig:generated images} shows examples of the generated MNIST and ImageNet images. For both figures, each column of images corresponds to a particular label class of the considered dataset. 

\subsection{Local Lipschitz Constant Estimation}\label{sec:exp-lip}
From Theorem \ref{thm:intrinsic robustness bound}, we observe that given a class of classifiers with risk at least $\alpha$, the derived intrinsic robustness upper bound is mainly decided by the perturbation strength $\epsilon$ and the local Lipschitz constant $L(r)$. While $\epsilon$ is usually predesignated in common robustness evaluation settings, the local Lipschitz constant $L(r)$ is unknown for most real world tasks. Computing an exact Lipschitz constant of a deep neural network is a difficult open problem. Thus, instead of obtaining the exact value, we approximate $L(r)$ using a sample-based approach with respect to the generative models. 

Recalling Definition \ref{def:lipschitz}, we consider $\Delta$ as the $\ell_2$ distance and  $g(\bz)$ and $g(\bz')$ are easy to compute via the generator network. Computing $L(r)$, however, is much more complicated as it requires obtaining a maximum value within a radius-$r$ ball. 
To deal with this, our approach approximates $L(r)$ by sampling $N$ points in the neighborhood around $\bz$ and takes the maximum value as the estimation of the true maximum value within the ball. Since the definition of local Lipschitz is probabilistic, we take multiple samples of the latent vectors $\bz$ to estimate the local Lipschitz constant $L(r)$.
The estimation procedure is summarized in Algorithm \ref{alg:lip}, which gives an underestimate of the underlying truth. 
Developing better Lipschitz estimation methods is an active area in machine learning research, but is not the main focus of this work.

\begin{algorithm}[htb]
	\caption{Local Lipschitz Estimation}
	\label{alg:lip}
	\begin{algorithmic}
		\STATE \textbf{Input:} number of samples $S$, number of local neighbors per sample $N$, $r$, $\delta$
 		\FOR {$i = 1,\ldots, S$}
 		    \STATE Generate a latent space sample $\bz_i$
		    \STATE Generate $N$ samples $\{\hat\bz_i^j\}_{j=1}^N$ within $\Ball_r(\bz_i)$ 
		    \STATE $L_i = \max_j \frac{\|g(\hat\bz_i^j) -g(\bz_i)\|_2}{\|\hat\bz_i^j-\bz_i\|_2}  $   
		\ENDFOR     
	    \STATE \textbf{Output:} $(1-\delta)$-percentile of $\{L_i\}_{i=1}^S$  
	\end{algorithmic}
\end{algorithm}

Tables~\ref{table:gan-lip} and \ref{table:gan-lip-imagenet} summarize the local Lipschitz constants estimated for the trained ACGAN and BigGAN generators conditioned on each class. In particular, we report both the mean estimates averaged over $10$ repeated trials and the standard deviations.
For both conditional generators, we set $S = 1000$, $N = 2000$, $r = 0.5$ and $\delta = 0.001$ in Algorithm \ref{alg:lip} for Lipschitz estimation.
For BigGAN, the specifically selected $10$ classes from ImageNet are reported in Table \ref{table:gan-lip-imagenet}. 

Compared with unconditional generative models, conditional ones generate each class using a separate generator. Thus, the local Lipschitz constant of each class-conditioned generator is expected to be smaller than that of unconditional ones, as the within-class variation is usually much smaller than the between-class variation for a given classification dataset.
For instance, we trained an unconditional GAN generator \citep{goodfellow2014generative} on MNIST dataset,
which yields an overall local Lipschitz constant of $27.01$ from Algorithm~\ref{alg:lip} 
under the same parameter settings. 
If we plug in this estimated Lipschitz constant into the theoretical results in \citet{fawzi2018adversarial}, the implied intrinsic robustness bound is in fact vacuous (above $1$) with perturbations strength $\epsilon\leq3.0$ in $\ell_2$ distance. 

\begin{table}[t] 
\caption{The estimated local Lipschitz constants of the trained ACGAN model on the $10$ MNIST classes with $r = 0.5$ and $\delta = 0.001$. }
\label{table:gan-lip}
  \centering
  \begin{tabular}{lccccc}
    \toprule
    Class & digit 0 & digit 1 & digit 2 & digit 3 & digit 4 \\
    \midrule
    Lipschitz & $7.9\pm 0.3$ & $8.6\pm 0.4$ & $8.3\pm 0.4$ & $7.8 \pm 0.3$ & $10.3 \pm 0.6$ \\
    \midrule
    \midrule
    Class & digit 5 & digit 6 & digit 7 & digit 8 & digit 9    \\
    \midrule
    Lipschitz & $11.0 \pm 0.4$ & $9.5\pm 0.3$ & $7.8 \pm 0.2$ & $9.3\pm 0.4$ & $10.9\pm 0.4$  \\
    \bottomrule
  \end{tabular}
\end{table}
 
\begin{table}[t] 
\caption{\normalsize The estimated local Lipschitz constants of the BigGAN model on the $10$ selected ImageNet classes with $r = 0.5$ and $\delta = 0.001$.}
  \label{table:gan-lip-imagenet}
  \centering
  \begin{tabular}{lccccc}
    \toprule
    Class & airliner & jeep & goldfinch & tabby cat & hartebeest \\
    \midrule
    Lipschitz & $13.1\pm 0.8$ & $14.5 \pm 1.1$ & $11.7\pm 0.5$ & $12.4\pm 0.4$ & $10.4\pm 1.1$  \\
    \midrule
    \midrule
    Class & Maltese dog & bullfrog & sorrel & pirate ship & pickup  \\
    \midrule
    Lipschitz & $11.3\pm 0.6$ & $9.4\pm 0.3$ & $13.0\pm 0.3$ & $13.1\pm 0.8$ & $14.9\pm 0.9$  \\
    \bottomrule
  \end{tabular}
\end{table}

\subsection{Comparisons with Robust Classifiers}
\label{sec:exp comp}

We compare our derived intrinsic robustness upper bound with the empirical adversarial robustness achieved by the current state-of-the-art defense methods under $\ell_2$ perturbations. 
Specifically, we consider three robust training methods: \textit{LP-Certify}: optimization-based certified robust defense \citep{wong2018scaling}; \textit{Adv-Train}: PGD attack based adversarial training  \citep{madry2017towards}; and \textit{TRADES}: adversarial training by accuracy and robustness trade-off \citep{zhang2019theoretically}. We adopt these robust training methods to train robust classifiers over a set of generated training images and evaluate their  robustness on the corresponding generated test set. 

For MNIST, we use our trained ACGAN model to generate $10$ classes of hand-written digits with $60,000$ training images and $10,000$ testing images. For ImageNet, we use the BigGAN model to generate $10$ selected classes of images, which contains $50,000$ images for training set and $10,000$ images for test set. We refer to the $10$-class BigGAN generated dataset as `ImageNet10'. 
We set $\epsilon=3.0$ for training robust models using Adv-Train and TRADES for both generated datasets, whereas we only train the LP-based certified robust classifier with $\epsilon=2.0$ on generated MNIST data, as it is not able to scale with ImageNet10 as well as generated MNIST with larger $\epsilon$ (see Appendix \ref{sec:cnn} for all the selected hyper-parameters and network architectures).

A commonly-used method to evaluate the robustness of a given model is by performing carefully-designed adversarial attacks. 
Here we adopt the PGD attack \citep{madry2017towards}, and report the robust accuracy (classification accuracy on inputs generated using the PGD attack) as the empirically measured model robustness. 
We test both the natural classification accuracy and the robustness of the aforementioned adversarially trained classifiers under $\ell_2$ perturbations with perturbation strength $\epsilon$ selected from $\{1.0, 2.0, 3.0\}$. See Appendix~\ref{sec:cnn} for PGD parameter settings.

\begin{table*}[h!]
  \caption{Comparisons between the empirically measured robustness of adversarially trained classifiers and the implied theoretical intrinsic robustness bound on the conditional generated datasets.}
  \label{table:acc}
  \small{
  \begin{center}
  \begin{tabular}{lccccc}
    \toprule
    \multirow{2}{*}[-2pt]{\textbf{Dataset}} & \multirow{2}{*}[-2pt]{\textbf{Method}} & \multirow{2}{*}[-2pt]{\textbf{Natural Accuracy}} & \multicolumn{3}{c}{\textbf{Adversarial Robustness}} \\
    \cmidrule(r){4-6}
    & & & $\epsilon=1.0$  & $\epsilon=2.0$  & $\epsilon=3.0$ \\
    \midrule 
    \multirow{4}{*}{\small
		\parbox{1.5cm}{Generated MNIST}} & LP-Certify & $88.3\pm 0.2\%$  & $74.0\pm 0.4\%$ & $51.1\pm 0.6\%$ & $23.5\pm 0.3\%$\\
    & Adv-Train  & $97.2\pm 0.2\%$ & $93.1 \pm 0.2\%$ & $83.5\pm 0.3\%$ & $58.9\pm 0.4\%$\\
    & TRADES & $98.3\pm 0.1\%$ & $94.8 \pm 0.2\%$ & $81.8\pm 0.4\%$ & $57.7\pm 0.4\%$\\
    & Our Bound & - &  $98.2\%$ & $97.8\%$ & $97.2\%$ \\
    \midrule 
    \multirow{3}{*}{\small ImageNet10} & Adv-Train & $82.1\pm 0.3\%$ & $67.8\pm 0.3\%$ & $47.1\pm 0.4\%$ & $23.4\pm 0.4\%$ \\
    & TRADES & $83.4\pm 0.3\%$ & $68.5\pm 0.3\%$ & $49.1\pm 0.5\%$ & $27.8\pm 0.5\%$ \\
    & Our Bound & - &  $83.5\%$ & $81.8\%$ & $80.0\%$ \\
    \bottomrule
  \end{tabular}
  \end{center}
  }
\end{table*}

Table \ref{table:acc} compares the empirically measured robustness of the trained robust classifiers and the derived theoretical upper bound on intrinsic robustness.  
For empirically measured adversarial robustness, we report both the mean and the standard deviation with respect to $10$ repeated trials.
For computing our theoretical robust bounds, we plug the estimated local Lipschitz constants into Theorem \ref{thm:intrinsic robustness bound} with risk threshold $\alpha=0.015$ for generated MNIST and $\alpha=0.15$ for ImageNet10, to reflect the best natural accuracy achieved by the considered robust classifiers. 

Under most settings, there exists a large gap between the robust limit implied by our theory and the best adversarial robustness achieved by state-of-the-art robust classifiers.
For instance, Adv-Train and TRADES only achieve less than $50\%$ robust accuracy on the generated ImageNet10 data with $\epsilon=2.0$, whereas the estimated robustness bound is as high as $81.8\%$. The gap becomes even larger when we increase the perturbation strength $\epsilon$.
In contrast to the previous theoretical results on artificial distributions, for these image classification problems we cannot simply conclude from the intrinsic robustness bound that adversarial examples are inevitable. This huge gap between the empirical robustness of the best current image classifiers and the estimated theoretical bound suggests that either there is a way to train better robust models or that there exist other explanations for the inherent limitations of robust learning against adversarial examples.

\subsection{In-distribution Adversarial Robustness}\label{sec:exp-rob}
In Section \ref{sec:exp comp}, we empirically show the unconstrained  robustness of existing robust classifiers is far below the intrinsic robustness upper bound implied by our theory for real distributions. 
However, it is not clear whether the reason is that current robust training methods are far from perfect, or that our derived upper bound is not tight enough due to the Lipschitz relaxation step used for proving such bound. In this section, we empirically study the in-distribution adversarial risk for a better characterization of the actual intrinsic robustness. As shown in Remark \ref{rem:relation}, the in-distribution adversarial robustness of any classifier with risk at least $\alpha$ can be regarded as a lower bound for the intrinsic robustness $\Rob_\mu^\epsilon(\cF_\alpha)$.
This provides us a more accurate characterization of the intrinsic robustness bound and enables better understanding of intrinsic robustness.

While there are many types of attack algorithms in the literature that can be used to evaluate the unconstrained robustness of a given classifier in the image space, little has been done in terms of how to evaluate the in-distribution robustness. 
In order to empirically evaluate the in-distribution robustness, we straightforwardly formulate the following optimization problem to find adversarial examples on the image manifold:
\begin{align}\label{eq:on-mani-att}
    \min_{\bz} \ \cL(f(G(\bz, y)), y) \quad \mbox{ s.t. } \:\: \|G(\bz, y) - \bx\|_2 &\leq \epsilon,
\end{align}
where $\bz \in \mathbb{R}^{d}$, $\bx$ is the data sample in the image space to be attacked, $f$ is the given classifier, and $\cL$ denotes the adversarial loss function. 
The goal of \eqref{eq:on-mani-att} is to optimize the latent vector to lower the adversarial loss (make the robust classifier mis-classify some generated images) while keeping the distance between the generated image and the test image within $\epsilon$ perturbation limit.
The key difficulty in solving \eqref{eq:on-mani-att} lies in the fact that we cannot perform any type of projection operations as we are optimizing over $\bz$ but the constraints are imposed on the generated image space $G(\bz, y)$. This prohibits the use of common attack algorithms such as PGD.  
In order to solve \eqref{eq:on-mani-att}, we transform \eqref{eq:on-mani-att} into the following Lagrangian formulation:
\begin{align}\label{eq:on-mani-final}
    \min_{\bz}  \|G(\bz, y) - \bx\|_2 + \lambda \cdot \cL(f(G(\bz, y)), y).
\end{align}
This formulation ignores the perturbation constraint of $\epsilon$ and tries to find the in-distribution adversarial examples with the smallest possible perturbation. In order to evaluate the intrinsic robustness under a given $\epsilon$ perturbation budget, we need to further check all in-distribution adversarial examples found and only count those with perturbations within the $\epsilon$ constraint. 
Note that even though \eqref{eq:on-mani-final} provides us a feasible way to compute the in-distribution robustness of a classifier, equation \eqref{eq:on-mani-final} itself could be hard to solve in general.
First, it is not obvious how to initialize $\bz$. Random initialization of $\bz$ could lead to bad local optima which prevent the optimizer from efficiently solving \eqref{eq:on-mani-final} or even finding a $\bz$ that could make $G(\bz, y)$ close enough to $\bx$. Second, the hyper-parameter $\lambda$ could be quite sensitive to different test examples. Failing to choose a proper $\lambda$ could also lead to failures in finding in-distribution adversarial examples within $\epsilon$ constraint. 
In order to the tackle the aforementioned challenges, we propose to solve another optimization problem for the initialization of $\bz$ and adopt binary search for the best choice of $\lambda$
(see Appendix \ref{sec:tricks} for more details of our implementation).

\begin{figure*}[t]
    \centering
    \subfigure[Generated MNIST ($\epsilon=1.0$)]{\includegraphics[width=0.315\textwidth]{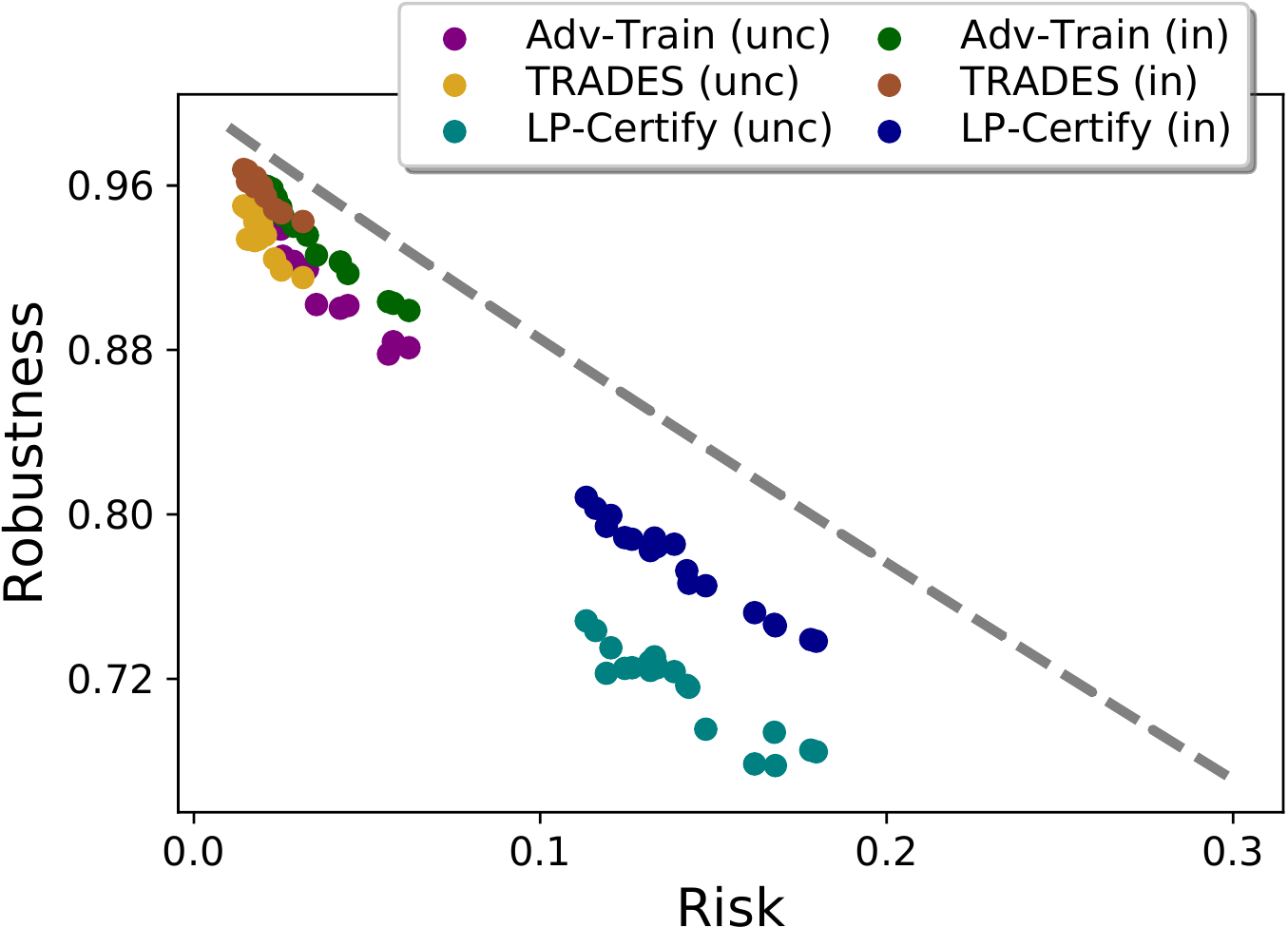}}
    \hspace{0.05in}
    \subfigure[Generated MNIST ($\epsilon=2.0$)]{\includegraphics[width=0.315\textwidth]{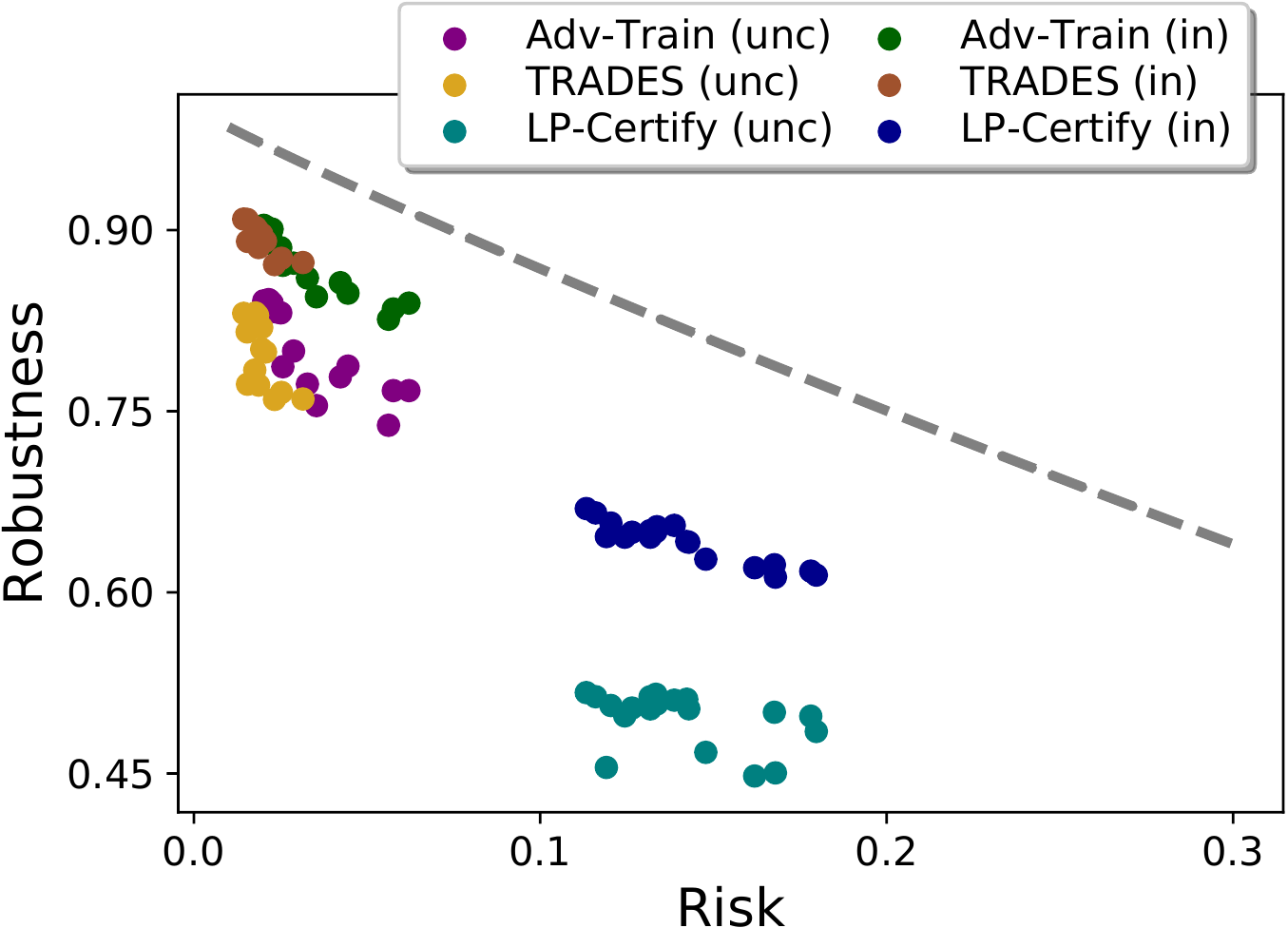}}
    \hspace{0.05in}
    \subfigure[Generated MNIST ($\epsilon=3.0$)]{\includegraphics[width=0.315\textwidth]{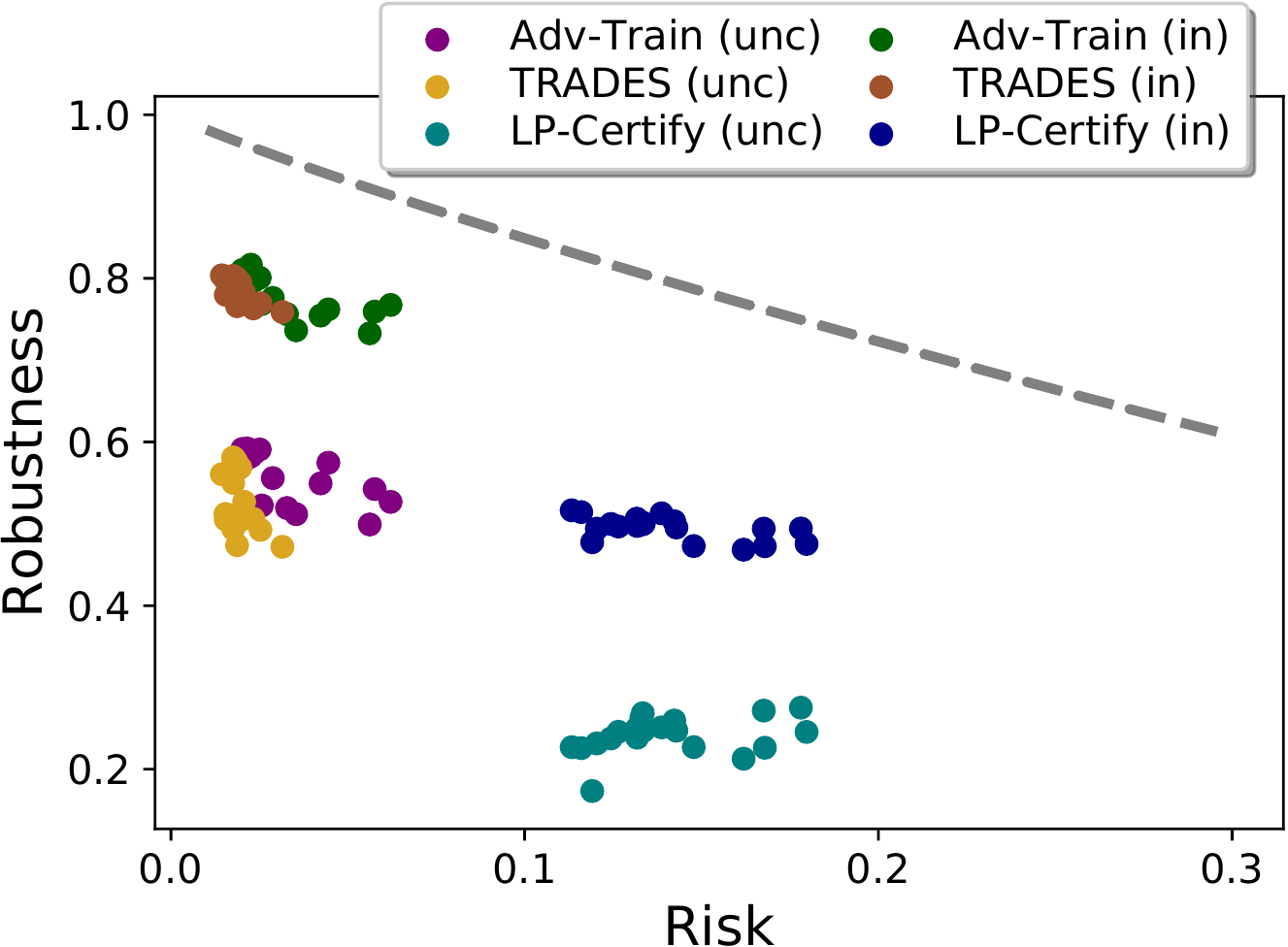}}
    \subfigure[ImageNet10 ($\epsilon=1.0$)]{\includegraphics[width=0.315\textwidth]{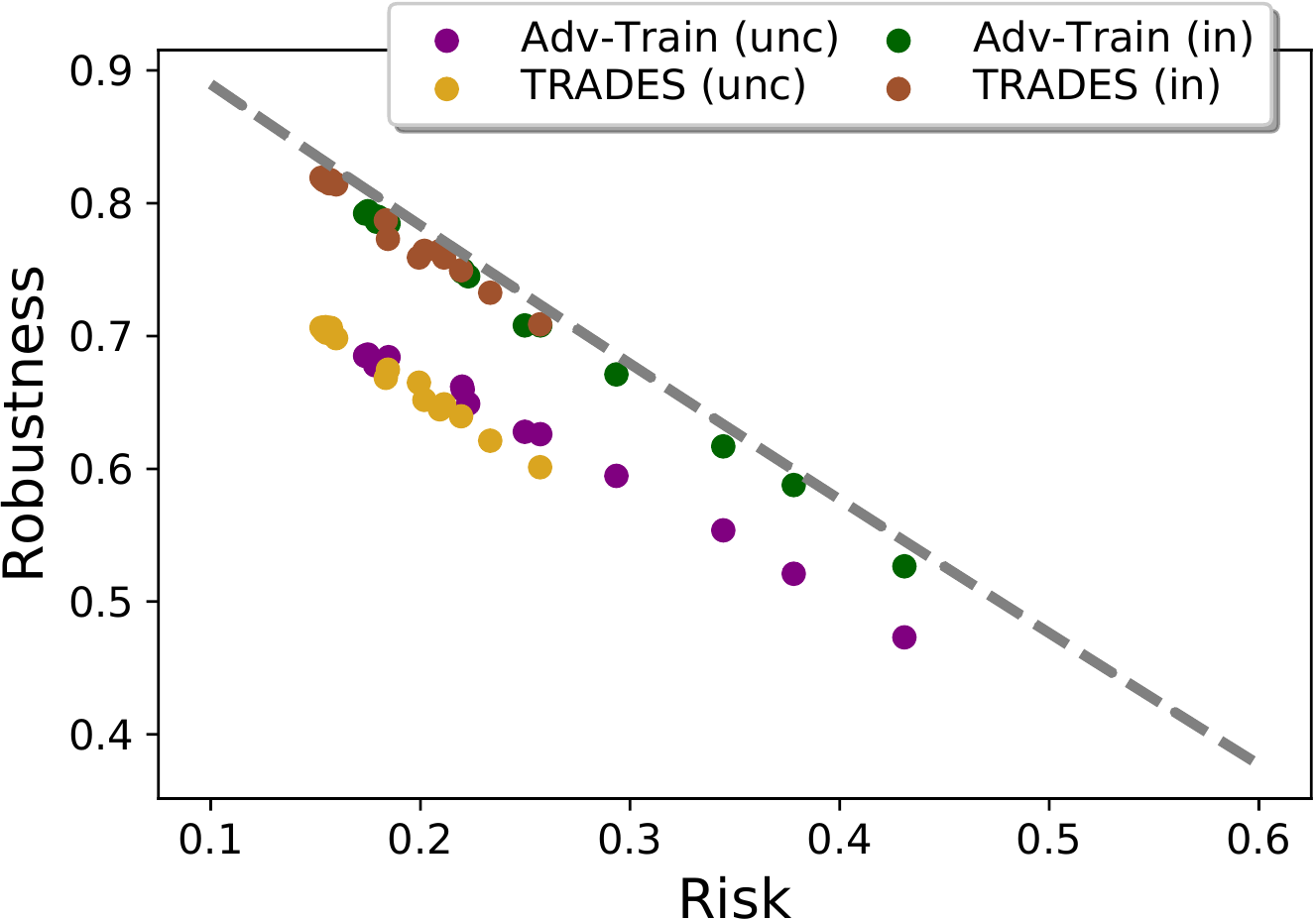}}
    \hspace{0.05in}
    \subfigure[ImageNet10 ($\epsilon=2.0$)]{\includegraphics[width=0.315\textwidth]{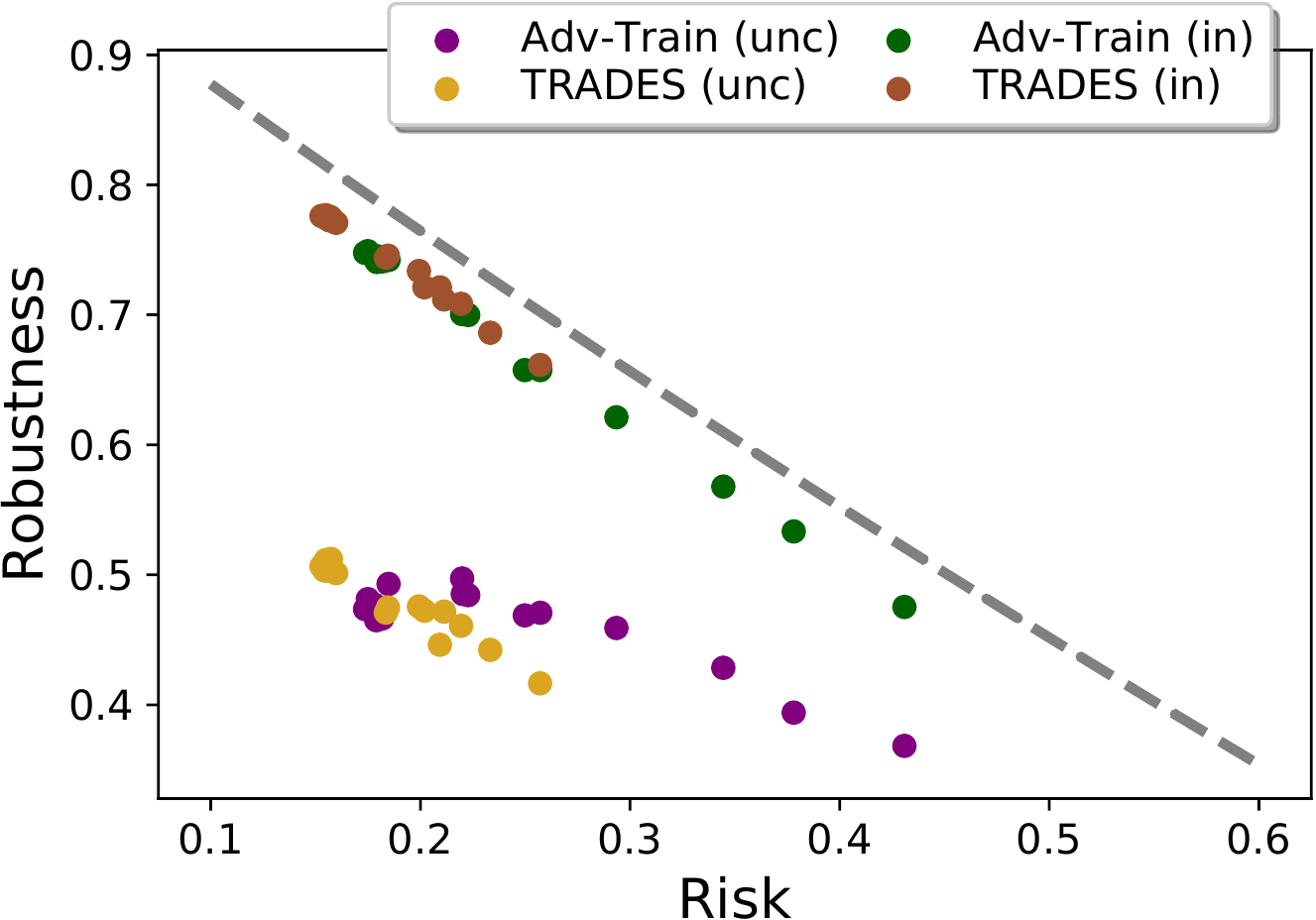}}
    \hspace{0.05in}
    \subfigure[ImageNet10 ($\epsilon=3.0$)]{\includegraphics[width=0.315\textwidth]{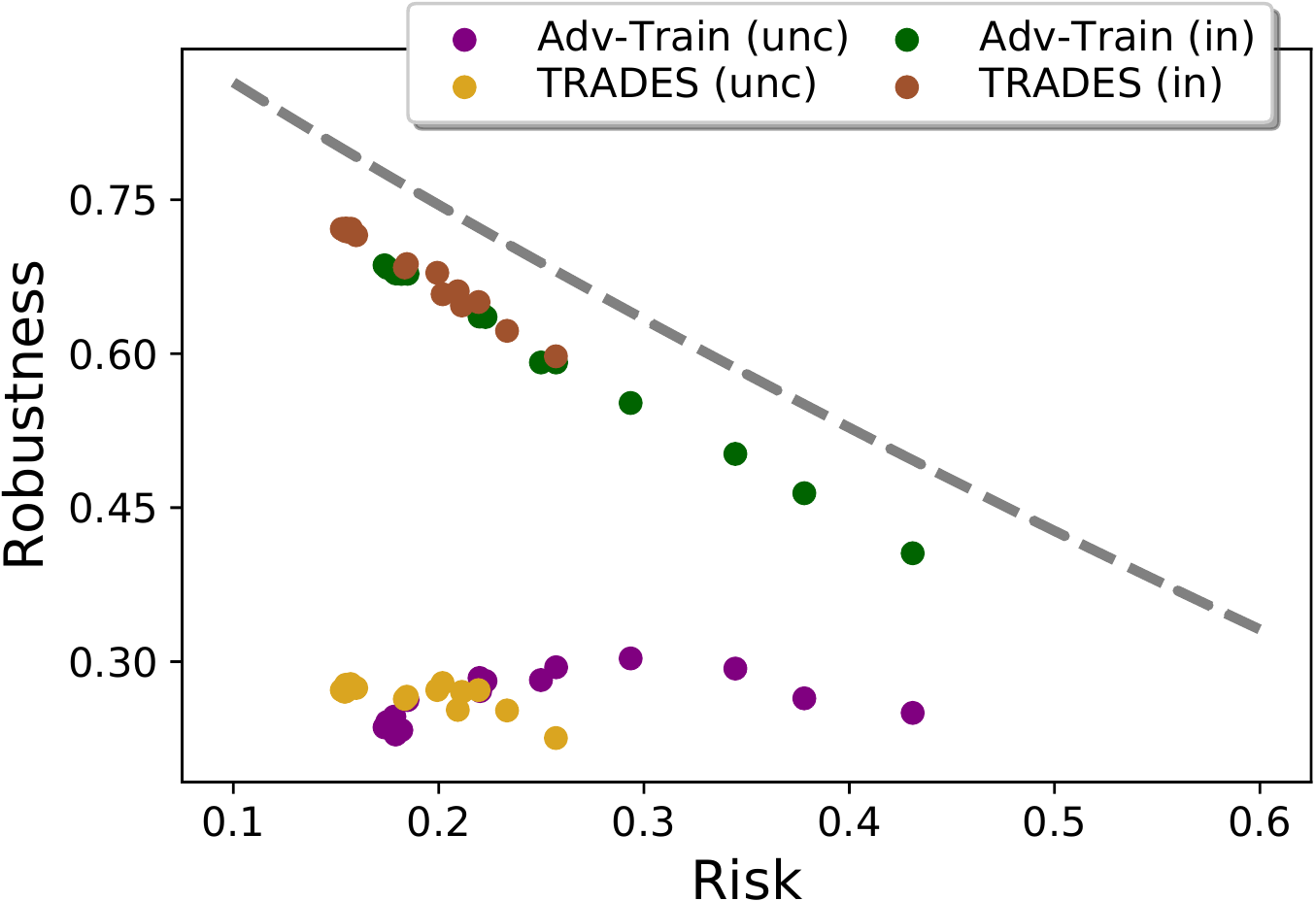}}
    \caption{Comparisons between the theoretical intrinsic robustness bound and the empirically estimated unconstrained/in-distribution adversarial robustness, denoted as ``unc'' and ``in'' in the legend, of models produced during robust training on the generated data under $\ell_2$. In each subfigure, the dotted curve line represents the theoretical bound on intrinsic robustness with horizontal axis denoting the different choice of $\alpha$.}
\label{fig:in dist rob}
\end{figure*}
 
Figure \ref{fig:in dist rob} summarizes results from our empirical evaluations on intrinsic robustness of the generated MNIST and ImageNet10 data. We evaluate the empirical robustness of three types of robust training methods at different time points during the training procedure. To be more specific, we evaluate the robustness of the intermediate models produced every $5$ training epochs.
For each method, we plot both the unconstrained robustness measured by PGD attacks and the in-distribution robustness measured using the aforementioned strategies. In addition, based on the local Lipschitz constants estimated in Section~\ref{sec:exp-lip},
we plot the implied theoretical bound on intrinsic robustness as the dotted line curve for direct comparison.

Compared with the intrinsic robustness upper bound (dotted curve line), the unconstrained robustness of various robustly-trained models is much smaller, and the gap between them becomes more obvious as we increase $\epsilon$. This aligns with our observations in Section \ref{sec:exp comp}.
However under all the considered settings, the estimated in-distribution adversarial robustness is much higher than the unconstrained one and closer to the theoretical upper bound, especially for the ImageNet10 data.
Note that according to Remark \ref{rem:relation}, the actual intrinsic robustness $\Rob_\mu^{\epsilon}(\cF_\alpha)$ should lie between the in-distribution robustness of any given classifier with risk at least $\alpha$ and the derived intrinsic robustness upper bound. 
Observing the big gap between the estimated in-distribution and unconstrained robustness of various robustly trained models, one would expect the current state-of-the-art robust models are still far from approaching the actual intrinsic robustness limit for real image distributions.

\section{Conclusions}
We studied the intrinsic robustness of typical image distributions using conditional generative models. By deriving theoretical upper bounds on intrinsic robustness and providing empirical estimates on the generated image distributions, we observed a large gap between the theoretical intrinsic robust limit and the best robustness achieved by state-of-the-art robust classifiers.
Our results imply that the inevitability of adversarial examples claimed in recent theoretical studies, such as \citet{fawzi2018adversarial}, do not apply to real image distributions, and suggest that there is a need for deeper understanding on the intrinsic robustness limitations for real data distributions.

\appendix
\section{Proof of Main Theorem}
This section presents the detailed proofs of Theorems \ref{thm:AdvRisk lower bound} and \ref{thm:intrinsic robustness bound} in Section \ref{sec:theory}.

\subsection{Proof of Theorem \ref{thm:AdvRisk lower bound}}
\label{sec:proof 1}

\begin{proof}
Let $\cE = \{\bx\in\cX: f(\bx)\neq f^*(\bx)\}$ be the error region in the image space and $\cE_\epsilon = \{\bx\in\cX: \Delta(\bx,\cE)\leq\epsilon\}$ be the $\epsilon$-expansion of $\cE$ in metric $\Delta$.
By Definition \ref{def:risk and adv risk}, we have
\begin{align*}
\AdvRisk_{\mu}^\epsilon(f) = \mu(\cE_\epsilon) = \sum_{i=1}^K p_i\cdot \mu_i(\cE_\epsilon) = \sum_{i=1}^K p_i\cdot \AdvRisk_{\mu_i}^\epsilon(f).
\end{align*}
Since according to Definition \ref{def:in dist adv risk}, we have $\AdvRisk_{\mu_i}^\epsilon(f)\geq \text{In-AdvRisk}_{\mu_i}^\epsilon(f)$ for any $i\in[K]$. Thus, it remains to lower bound each term $\text{In-AdvRisk}_{\mu_i}^\epsilon(f)$ individually. For any classifier $f$, we have
\begin{align}
\label{eq:individual AdvRisk bound}
\nonumber\text{In-AdvRisk}_{\mu_i}^\epsilon(f) &= \Pr_{\bz\sim\nu_d} \Big[\exists\: \bz'\in\RR^d, \text{ s.t. } \Delta\big(g_i(\bz'),g_i(\bz)\big)\leq\epsilon \text{ and } f\big(g_i(\bz')\big)\neq f^*\big(g_i(\bz')\big)\Big] \\
&\geq \underbrace{\Pr_{\bz\sim\nu_d}\Big[
\exists\: \bz'\in\Ball\big(\bz, \epsilon/L_i(r)\big), \text{ s.t. } f\big(g_i(\bz')\big)\neq f^*\big(g_i(\bz')\big)\Big]}_{I} - \delta 
\end{align}
where the first inequality is due to $\mu_i=(g_i)_*(\nu_d)$, and the second inequality holds because $g_i$ is $L_i(r)$-locally Lipschitz with probability at least $1-\delta$ and $\Ball\big(\bz, \epsilon/L_i(r)\big)\subseteq\Ball\big(\bz, r\big)$ for any $\bz\in\RR^d$. 

To further bound the term $I$, we make use of the Gaussian Isoperimetric Inequality as presented in Lemma \ref{lem:Gaussian isoperimetric inq}. Let $\cA_f = \{\bz\in\RR^d: f(g_i(\bz))\neq f^*(g_i(\bz))\}$ be the corresponding error region in the latent space. By Lemma \ref{lem:Gaussian isoperimetric inq}, we have
\begin{align}
\label{eq:latent space bound}
I &\geq \Phi\bigg(\Phi^{-1}\big(\nu_d(\cA_f)\big)+\frac{\epsilon}{L_i(r)}\bigg) =  \Phi\bigg(\Phi^{-1}\big(\Risk_{\mu_i}(f)\big)+\frac{\epsilon}{L_i(r)}\bigg).
\end{align}
Finally, plugging \eqref{eq:latent space bound} into \eqref{eq:individual AdvRisk bound}, we complete the proof.
\end{proof}

\subsection{Proof of Theorem \ref{thm:intrinsic robustness bound}}
\label{sec:proof 2}
\begin{proof}
According to Definition \ref{def:intrinsic robustness} and Theorem \ref{thm:AdvRisk lower bound}, for any $f\in\cF_\alpha$, we have
\begin{align}
\label{eq:intrinisic robustness bound}
\nonumber\Rob_\mu^\epsilon(\cF_\alpha)
&\leq 1+\delta - \sum_{i=1}^K p_i\cdot\Phi\bigg(\Phi^{-1}\big(\Risk_{\mu_i}(f)\big)+ \frac{\epsilon}{L_i(r)} \bigg) \\
    &\leq 
    1+\delta - \sum_{i=1}^K p_i\cdot\Phi\bigg(\Phi^{-1}\big(\Risk_{\mu_i}(f)\big)+ \frac{\epsilon}{L_{\max}(r)} \bigg),
\end{align}
where the last inequality holds because $\Phi(\cdot)$ is monotonically increasing. For any $f\in\cF_\alpha$, let $\cE = \{\bx\in\cX: f(\bx)\neq f^*(\bx)\}$ be the error region and $\alpha_i = \mu_i(\cE)$ be the measure of $\cE$ under the $i$-th conditional distribution. 

Thus, to obtain an upper bound on $\Rob_\mu^\epsilon(\cF_\alpha)$ using \eqref{eq:intrinisic robustness bound}, it remains to solve the following optimization problem:
\begin{align}
\label{eq:bound opt}
    \minimize_{\alpha_1,\ldots,\alpha_K\in[0,1]}\: \sum_{i=1}^K p_i\cdot\Phi\bigg(\Phi^{-1}(\alpha_i)+ \frac{\epsilon}{L_{\max}(r)} \bigg) \quad \text{subject to}\:\: \sum_{i=1}^K p_i\alpha_i \geq \alpha.
\end{align}
Note that for classifier in $\tilde\cF_\alpha$, by definition, we can simply replace $\alpha_i=\alpha$ in \eqref{eq:bound opt}, which proves the upper bound on $\Rob_\mu^\epsilon(\tilde\cF_\alpha)$.

Next, we are going to show that the optimal value of \eqref{eq:bound opt} is achieved, only if there exists a class $i'\in[K]$ such that $\alpha_{i'} = \alpha/p_{i'}$ and $\alpha_i=0$ for any $i\neq i'$. Consider the simplest case where $K=2$. 
Note that $\Phi(\cdot)$ and $\Phi^{-1}(\cdot)$ are both monotonically increasing functions, which implies that $\sum_{i=1}^{K} p_i\alpha_i = \alpha$ holds when optimum achieved, thus the optimization problem for $K=2$ can be formulated as follows
\begin{align}
\label{eq:opt K=2}                 \min_{\alpha_1,\alpha_2\in[0,1]}\:\:  p_1\cdot\Phi\bigg(\Phi^{-1}(\alpha_1)+ \frac{\epsilon}{L_{\max}(r)} \bigg) + p_2\cdot\Phi\bigg(\Phi^{-1}(\alpha_2)+ \frac{\epsilon}{L_{\max}(r)} \bigg) \quad \text{s.t.}\:\: p_1\alpha_1+p_2\alpha_2 = \alpha.
\end{align}

Suppose $\alpha_1 \geq \alpha_2$ holds for the initial setting. Now consider another setting where $\alpha_1'>\alpha_1$, $\alpha_2'<\alpha_2$. Let $s_1 = \Phi^{-1}(\alpha_1')-\Phi^{-1}(\alpha_1)$ and $s_2 = \Phi^{-1}(\alpha_2)-\Phi^{-1}(\alpha_2')$. According to the equality constraint of the optimization problem \eqref{eq:opt K=2}, we have
\begin{align}
\label{eq:equality constraint}
p_1\cdot\int_{\Phi^{-1}(\alpha_1)}^{\Phi^{-1}(\alpha_1)+s_1} \frac{1}{\sqrt{2\pi}}\cdot\exp^{-x^2/2}dx = p_2\cdot\int_{\Phi^{-1}(\alpha_2)-s_2}^{\Phi^{-1}(\alpha_2)} \frac{1}{\sqrt{2\pi}}\cdot\exp^{-x^2/2}dx.
\end{align}
Let $\eta = \epsilon/L_{\max}(r)$ for simplicity. By simple algebra, we have
\begin{align*}
p_1\cdot\int_{\Phi^{-1}(\alpha_1)+\eta}^{\Phi^{-1}(\alpha_1)+s_1+\eta} \frac{1}{\sqrt{2\pi}}\cdot\exp^{-x^2/2}dx &= p_1\cdot\int_{\Phi^{-1}(\alpha_1)}^{\Phi^{-1}(\alpha_1)+s_1} \frac{1}{\sqrt{2\pi}}\cdot\exp^{-u^2/2-\eta\cdot u-\eta^2/2}du \\
&< p_1\cdot\exp^{-\eta\cdot \Phi^{-1}(\alpha_1)-\eta^2/2}\cdot\int_{\Phi^{-1}(\alpha_1)}^{\Phi^{-1}(\alpha_1)+s_1} \frac{1}{\sqrt{2\pi}}\cdot\exp^{-u^2/2}du \\
&\leq p_2\cdot\exp^{-\eta\cdot \Phi^{-1}(\alpha_2)-\eta^2/2}\cdot\int_{\Phi^{-1}(\alpha_2)-s_2}^{\Phi^{-1}(\alpha_2)} \frac{1}{\sqrt{2\pi}}\cdot\exp^{-u^2/2}du \\
&< p_2\cdot\int_{\Phi^{-1}(\alpha_2)-s_2+\eta}^{\Phi^{-1}(\alpha_2)+\eta} \frac{1}{\sqrt{2\pi}}\cdot\exp^{-x^2/2}dx,
\end{align*}
where the first inequality holds because $\exp^{-\eta\cdot u} < \exp^{-\eta\cdot \Phi^{-1}(\alpha_1)}$ for any $u>\Phi^{-1}(\alpha_1)$, the second inequality follows from \eqref{eq:equality constraint} and the fact that $\Phi^{-1}(\alpha_1)\geq\Phi^{-1}(\alpha_2)$, and the last inequality holds because $\exp^{-\eta\cdot\Phi^{-1}(\alpha_2)}<\exp^{-\eta\cdot u}$ for any $u<\Phi^{-1}(\alpha_2)$. Therefore, the optimal value of \eqref{eq:opt K=2} will be achieved when $\alpha_1=0$ or $\alpha_2=0$. For general setting with $K>2$, since $\alpha_1,\ldots,\alpha_K$ are independent in the objective, we can fix $\alpha_3,\ldots,\alpha_K$ and optimize $\alpha_1$ and $\alpha_2$ first, then deal with $\alpha_i$ incrementally using the same technique.
\end{proof}


\section{Experimental Details}
This section provides additional details for our experiments.

\subsection{Network Architectures and Hyper-parameter Settings}
\label{sec:cnn}
For the certified robust defense (LP-Certify), we adopt the the same four-layer neural network architecture as  implemented in \citet{wong2018scaling}, with two convolutional layers and two fully connected layers, and use the an Adam optimizer with learning rate $0.001$ and batch size $50$ for training the robust classifier. In particular, the adversarial loss function is based on the robust certificate under $\ell_2$ proposed in \citet{wong2018scaling}.

For training attack-based robust models (Adv-Train and TRADES), we use a seven-layer CNN architecture which contains four convolution layers and three fully connected layers.
We use a SGD optimizer to minimize the attack-based adversarial loss with learning rate $0.05$ on MNIST and learning rate $0.01$ on ImageNet10. Table \ref{table:para_train} summarizes all the hyper-parameters we used for training the robust models ($\beta$ is an additional parameter specifically used in TRADES).

For evaluating the unconstrained adversarial robustness, we implemented PGD attack with $\ell_2$ metric.
Table \ref{table:para_test} shows all the hyper-parameters we used for robustness evaluation.

\begin{table}[h!]
  \caption{Hyper-parameters used for training robust models.
  }
  \label{table:para_train}
  \begin{center}
  \begin{small}
  \begin{tabular}{l|ccc|cc}
    \toprule
    \multicolumn{1}{c}{\multirow{2}{*}{Para.}} & \multicolumn{3}{c}{Generated MNIST} &
    \multicolumn{2}{c}{ImageNet10}\\
    \cmidrule(r){2-4}
    \cmidrule(r){5-6}
    \multicolumn{1}{c}{} & LP-Certified & Adv Training & TRADES & Adv Training & TRADES \\
    \midrule 
    $\epsilon$ (in $\ell_2$) & $2.0$  & $3.0$ & $3.0$ & $3.0$ & $3.0$\\
    optimizer & ADAM  & SGD & SGD & SGD & SGD\\
    learning rate & $0.001$  & $0.05$ & $0.05$ & $0.01$ & $0.01$\\
    \#epochs & $60$ & $100$ & $100$ & $100$ & $100$ \\
    attack step size & - & $0.5$  & $0.5$  & $0.5$  & $0.5$ \\
    \#attack steps & - & $40$ & $40$ & $10$ & $10$ \\
    $\beta$ & - & - & $6.0$ & - & $6.0$ \\
    \bottomrule
  \end{tabular}
  \end{small}
  \end{center}
\end{table}

\begin{table}[h!]
  \caption{Hyper-parameters used for evaluating the model robustness via PGD attack.
  }
  \label{table:para_test}
  \begin{center}
  \begin{tabular}{l|ccc|ccc}
    \toprule
    \multicolumn{1}{c}{\multirow{2}{*}{Para.}} & \multicolumn{3}{c}{Generated MNIST} &
    \multicolumn{3}{c}{ImageNet10}\\
    \cmidrule(r){2-4}
    \cmidrule(r){5-7}
    \multicolumn{1}{c}{} & $\epsilon=1.0$ & $\epsilon=2.0$ & $\epsilon=3.0$ & $\epsilon=1.0$ & $\epsilon=2.0$ & $\epsilon=3.0$ \\
    \midrule 
    attack step size  & $0.1$  & $0.3$ & $0.5$ & $0.1$ & $0.3$ & $0.5$ \\
    \#attack steps & $100$ & $100$ & $100$ & $100$ & $100$ & $100$\\
    \bottomrule
  \end{tabular}
  \end{center}
\end{table}

\subsection{Strategies for Estimating In-distribution Adversarial Robustness}\label{sec:tricks}

\textbf{Initialization of $\bz$}: 
For MNIST data, we design an initialization strategy for $\bz$ in order to make sure the perturbation term $\|G(\bz, y) - \bx\|_2$ can be efficiently optimized. To be more specific, starting from random noise, we first solve another optimization problem: 
\begin{align*}
    \bz_{\text{init}} = \argmin_{\bz}  \|G(\bz, y) - \bx\|_2 .
\end{align*}
By setting $\bz_{\text{init}}$ as our initial point, we minimize the initial perturbation distance. Here $\bz$ can start from any random initial point as we will then optimize the generated image under $\ell_2$ distance.

For ImageNet10 data, even applying the above optimization procedure doesn't result in an initial $\bz$ such that $\|G(\bz, y) - \bx\|_2 \leq \epsilon$ when $\epsilon$ is small. Therefore, we use another strategy by recording the $\bz^*$ when generating the test sample $\xb$, i.e., $G(\bz^*, y) = \xb$. And we adopt $\bz^*$ as the initial point for $\bz$ in solving \eqref{eq:on-mani-final}. This makes sure that the whole optimization procedure could at least find one point satisfying the perturbation constraint\footnote{We didn't use $\bz^*$ as the initialization for MNIST data as our empirical study shows that the optimization-based initialization achieves better performances on MNIST.}.

\textbf{The choice of $\lambda$}:
Inspired by \cite{carlini2017towards}, we also adopt binary search strategy for finding better regularization parameter $\lambda$. Specifically, we set initial $\lambda = 1.0$ and if we successfully find an adversarial example, we lower the value of $\lambda$ via binary search. Otherwise, we raise the value of $\lambda$. For each batch of examples, we perform $5$ times binary search in order to find qualified in-distribution adversarial examples. 

\textbf{Hyper-parameters}: We use Adam optimizer with learning rate $0.01$ for finding in-distribution adversarial examples. We set maximum iterations for each $\lambda$ binary search as $10000$.

\subsection*{Acknowledgements}
This research was sponsored in part by the National Science Foundation SaTC-1717950 and SaTC-1804603, and additional support from Amazon, Baidu, and Intel.
The views and conclusions contained in this paper are those of the authors and should not be interpreted as representing any funding agencies.

\bibliographystyle{apalike}
\bibliography{ref}

\end{document}